\documentclass{ifacconf}

\usepackage{graphicx}      
\usepackage[square,numbers]{natbib}
\newcommand{\capequation}[1]{\begin{center} #1 \end{center}}
\usepackage{amssymb, amsmath}

\begin{document}
\begin{frontmatter}

\title{Online Learning Based Mobile Robot Controller Adaptation for Slip Reduction} 

\author[First]{Huidong Gao} 
\author[First]{Rui Zhou} 
\author[First]{Masayoshi Tomizuka}
\author[First]{Zhuo Xu}

\address[First]{Department of Mechanical Engineering, University of California, Berkeley, CA 94720 USA (e-mail: \{hgao9, ruizhouzr, tomizuka, zhuoxu\}@berkeley.edu)}

\begin{abstract}                

Slip is a very common phenomena present in wheeled mobile robotic systems. It has undesirable consequences such as wasting energy and impeding system stability. To tackle the challenge of mobile robot trajectory tracking under slippery conditions, we propose a hierarchical framework that learns and adapts gains of the tracking controllers simultaneously online. Concretely, a reinforcement learning (RL) module is used to auto-tune parameters in a lateral predictive controller and a longitudinal speed PID controller. Experiments show the necessity of simultaneous gain tuning, and have demonstrated that our online framework outperforms the best baseline controller using fixed gains. By utilizing online gain adaptation, our framework achieves robust tracking performance by rejecting slip and reducing tracking errors when the mobile robot travels through various terrains.

\end{abstract}


\end{frontmatter}

\section{Introduction}

\subsection{Background and Motivation}

 Mobile robots are used in various industrial applications such as manufacturing, process and aerospace. They often run on slippery terrains, or routes with rapid cornering, which induces skidding and slipping. Excessive slip may cause motion instability, undermine maneuverability and lead to possible collisions, thus should be prevented. 

 To mitigate slip, many works try to identify slip parameters or terrain states, and design simple control laws with robot kinematic models, e.g. \cite{pico2022climbing} and \cite{kim2016kinematic}.  \cite{sebastian2019active} and \cite{wang2020trajectory} further choose to model slip as disturbance in the kinematics model and estimate by observers. However, state vectors are of high order and matrix inverse calculations could be massive. Another line of work focuses on wheel dynamics with traction forces. \cite{tian2009modeling} choose to use the Magic formula to derive the relationship between traction force and slip ratio, while \cite{nandy2011detailed} formulates a detailed slip dynamics with certain switching conditions. Although dynamic models consider forces in addition to kinematics models, they usually require system identification for different scenarios and have poor generalization abilities. 

There are also other works utilizing reinforcement learning to directly learn a policy; such as in \cite{xu2018zero, tang2019disturbance, chang2020cascade, cai2020high, xu2021cocoi}. However, end-to-end RL approaches require considerable training time and pose challenges in explainability. Instead of end-to-end RL, \cite{carlucho2019double, gao2022reinforcement} uses RL only to optimize controllers, but the results are highly dependent on action space discretization.

\subsection{Contributions}

Our work focuses on trajectory tracking control of mobile robots under slippery conditions. We follow a similar approach as in \cite{carlucho2019double}, and propose to use a hierarchical framework that optimizes gains for the tracking controllers online. An RL module is used to tune gains in a lateral predictive Stanley controller and a longitudinal speed PID controller simultaneously for regulation. By dividing the control part into longitudinal and lateral control modules and tuning gains directly, we are able to improve lateral and speed tracking errors in a straight-forward way. By using a higher level RL module, we are able to tune multiple low-level controllers simultaneously in real-time. Furthermore, the RL module is only able to determine the conservativeness in the controllers, and thus the entire framework is more explainable than an end-to-end RL controller.

The contributions of our work can be summarized as follows: 1) We propose an hierarchical framework that actively optimizes controllers to slip conditions through RL gain-tuning. 2) We reason the necessity of simultaneous online gain tuning through experiments. 3) We demonstrate that our adaptive framework outperforms the best fixed-gain baselines by 6.6\% and 12.7\% for average lateral error and max lateral error by simulation.



\section{Methodology}
\subsection{Problem Overview}

\begin{figure}[ht]
\centering
\includegraphics[width=1.0\linewidth]{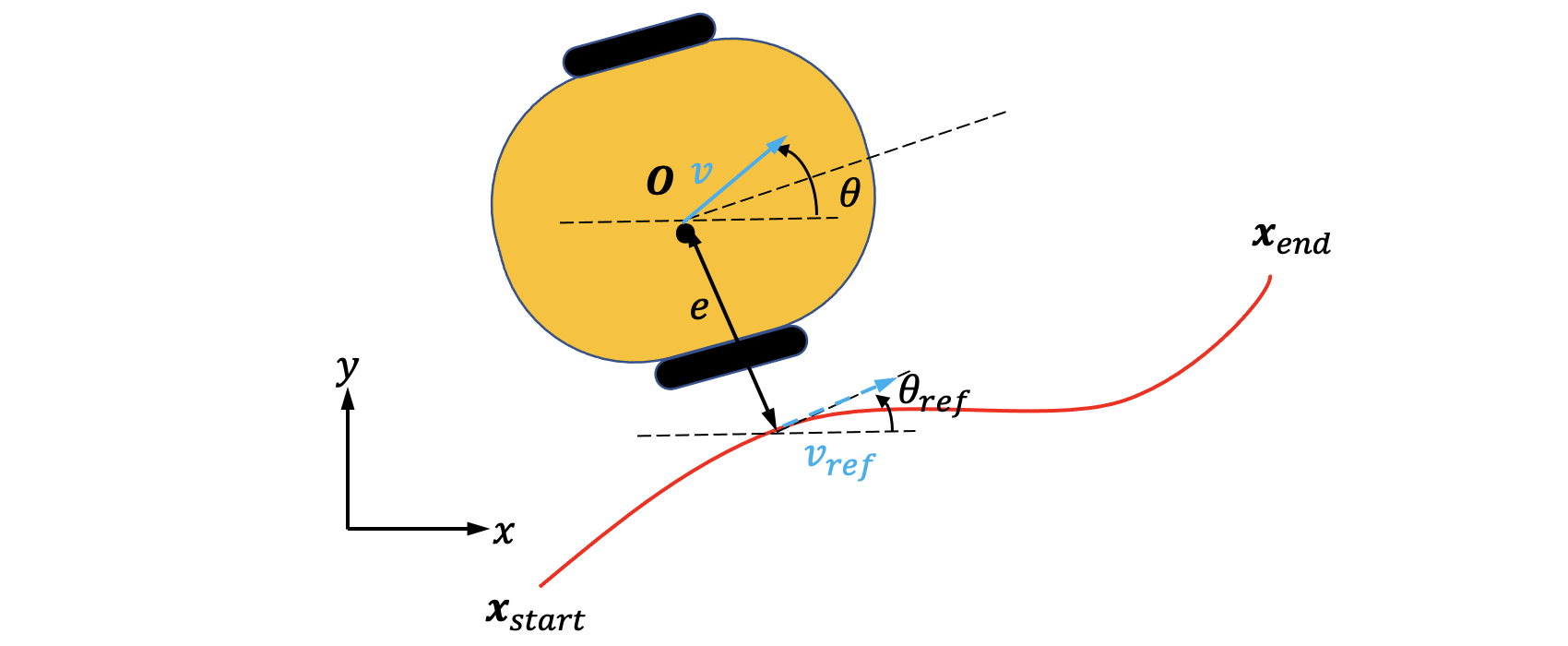}%
\caption{\textbf{Schematic diagram of the problem setup.} }
\label{fig:overview}
\end{figure}

Fig. \ref{fig:overview} illustrates our tracking problem layout. The robot's goal is to travel from $x_{start}$, following a predefined trajectory to reach $x_{end}$. Here we define certain terms to describe the robot's motion and the tracking state.  We use lateral displacement error $e$ to represent the closest tracking error relative to the reference trajectory (in unit $m$). $\Delta \theta$ is the yaw error, which is the difference between reference yaw $\theta_{ref}$ and actual yaw $\theta$ (in unit $rad$). $\Delta v$ is the speed error, which is the difference between absolute values of reference velocity $v_{ref}$ and actual velocity $v$ (in unit $m/s$).

The tracking task can then be formulated as a Markov Decision Process defined by $\mathcal{M} = (\mathcal{S},\mathcal{A}, \mathcal{T}, \mathcal{R},\gamma)$. $\mathcal{S}$ represents the
state space; $\mathcal{A}$ is the action space; $\mathcal{T}(s^{'}|s,a)$ is the state
transition model; $\mathcal{R}(s,a)$ is the reward function; and $\gamma \in [0,1)$ is the discount factor. The RL formulation aims to learn a policy $\pi(a|S)$.
The agent then follows the policy $\pi$, obtains an observation $s_t$ at time $t$ and performs an action
$a_t$. It then receives from the environment a reward $R_t$ and a new observation $s_{t+1}$, and $\pi$ is updated accordingly. The final trained model gives an action selection policy $\pi$ that maximizes the expectation of a discounted sum of rewards $E[\sum_{t=1}^{T}\gamma^{t-1}R_t]$.

The state $S$ has 5 variables: $e$, $\Delta\theta$, $\Delta v$, $\Delta v_{c\_vs\_actual}$, and $\Delta \omega_{c\_vs\_actual}$. 
$e$, $\Delta\theta$, and $\Delta v$ are as discussed in the beginning of the section. $\Delta v_{c\_vs\_actual}$ (in unit $m/s$)and $\Delta \omega_{c\_vs\_actual}$ (in unit $rad/s$) represent the difference between actual body velocities and body velocities calculated from wheel velocity commands(shown in Eqn. \ref{eqn:v_transformation} and \ref{eqn:w_transformation}). Intuitively, a large value of $\Delta v_{c\_vs\_actual}$ or $\Delta \omega_{c\_vs\_actual}$ indicates the robot is slipping more severely as the wheel velocity commands are not fully transferred to actual body velocities.

\vspace{1mm}

\begin{equation}
    v = \frac{(\omega_R+\omega_L) \cdot R}{2}
    \label{eqn:v_transformation}
\end{equation}

\begin{equation}
    \omega = \frac{(\omega_R-\omega_L) \cdot R}{b}
    \label{eqn:w_transformation}
\end{equation}
\capequation{Eqn. 1-2: transformation from wheel commands to calculated body linear and angular velocities. $\omega_R$ and $\omega_L$ are right and left wheel angular velocity commands, $R$ is wheel radius, and $b$ is distance between wheels. }

The action $A$ is [$v$, $\omega$], which are linear and angular velocity commands. Notice the final input wheel velocity commands are calculated from reversing equations 1-2. The low level controller executes the commands and gets a reward at this step. The reward for each step is defined in Eqn. \ref{eqreward}. Here we penalize $e$, $\Delta\theta$ and $\Delta v$, with coefficients $R_{dist}$, $R_{ang}$ and $R_{speed}$. The cumulative reward is defined as $\sum_{t=1}^{T}\gamma^{t-1}R_t$, where $R_t$ is the step reward.
\begin{equation}
R_t(s_t,a_t) = R_{dist} \cdot e^2 + R_{ang} \cdot \Delta\theta^2 + R_{speed} \cdot  \Delta v^2
\label{eqreward}
\end{equation}

\subsection{Proposed Framework}

\begin{figure}[ht]
\centering
\includegraphics[width=0.9\linewidth]{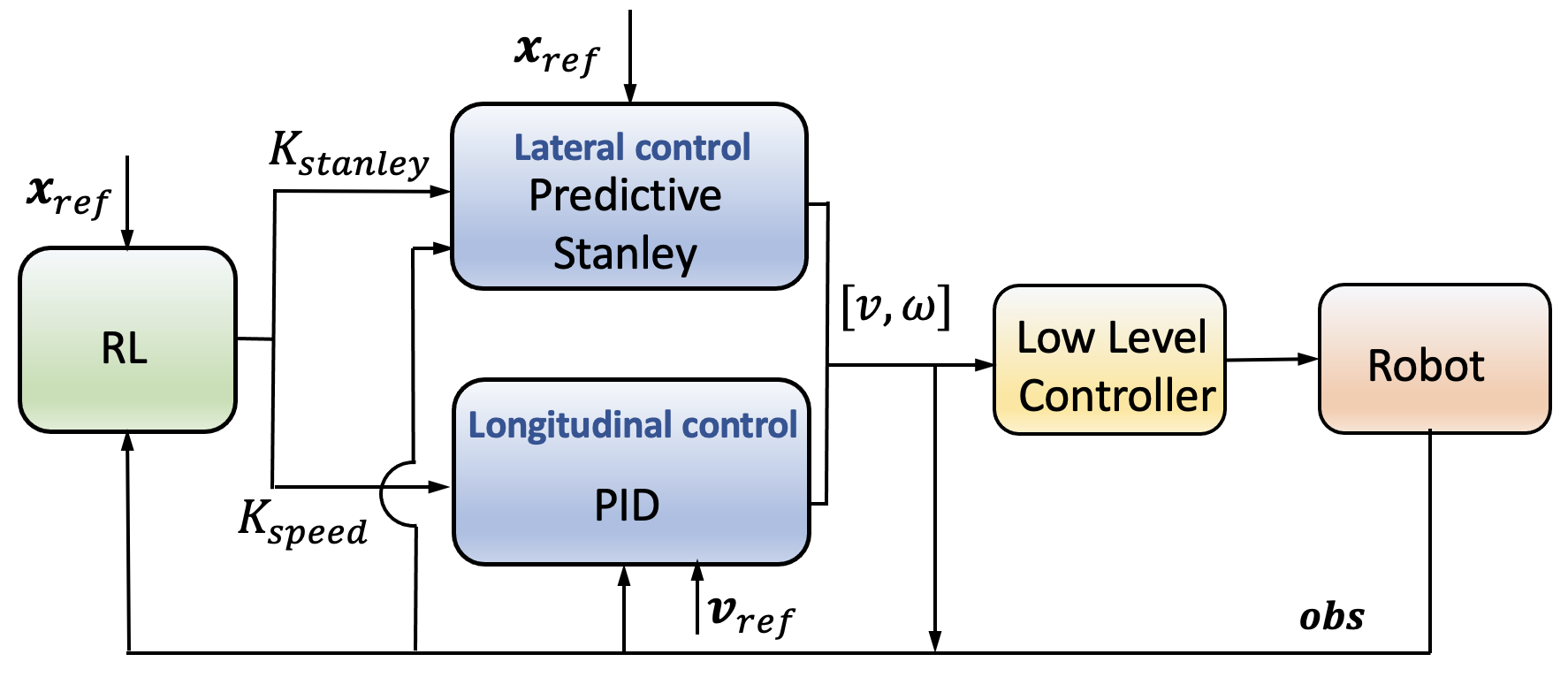}%
\caption{\textbf{Proposed framework.}}
\label{fig:framework}
\end{figure}

We propose to utilize reinforcement learning to actively tune parameters in lateral and longitudinal control modules on a differential drive TurtleBot. The proposed framework consists of a RL-based high-level module, a lateral control module, a longitudinal control module,  a low-level tracking controller, and the robot. The framework is visualized in Fig. \ref{fig:framework}. The RL module takes observed robot states $obs$; reference trajectory $x_{ref}$, and outputs gains $K_{stanley}$ and $K_{speed}$. The two control modules use the gains accordingly and output acceleration command $\alpha$ and steering angle command $\delta$, and then transfer them into linear and angular velocity commands $[v, \omega]$. The low level controller then executes the command on the robot and feeds the directly observed states back into the RL module to calculate for step rewards, and the policy is updated accordingly. The loop stops when the positional error of the robot and final goal position is within a threshold. The entire framework realizes our MDP formulation, and is trained end-to-end using an RL algorithm. 


\subsection{Lateral Control: Predictive Stanley}
We tackle our trajectory tracking problem by breaking it up into longitudinal and lateral control problems. The longitudinal controller is responsible for regulating the robot's speed while the lateral controller aims to reduce the lateral error during path tracking.

\begin{figure}[ht]
\centering
\includegraphics[width=0.8\linewidth]{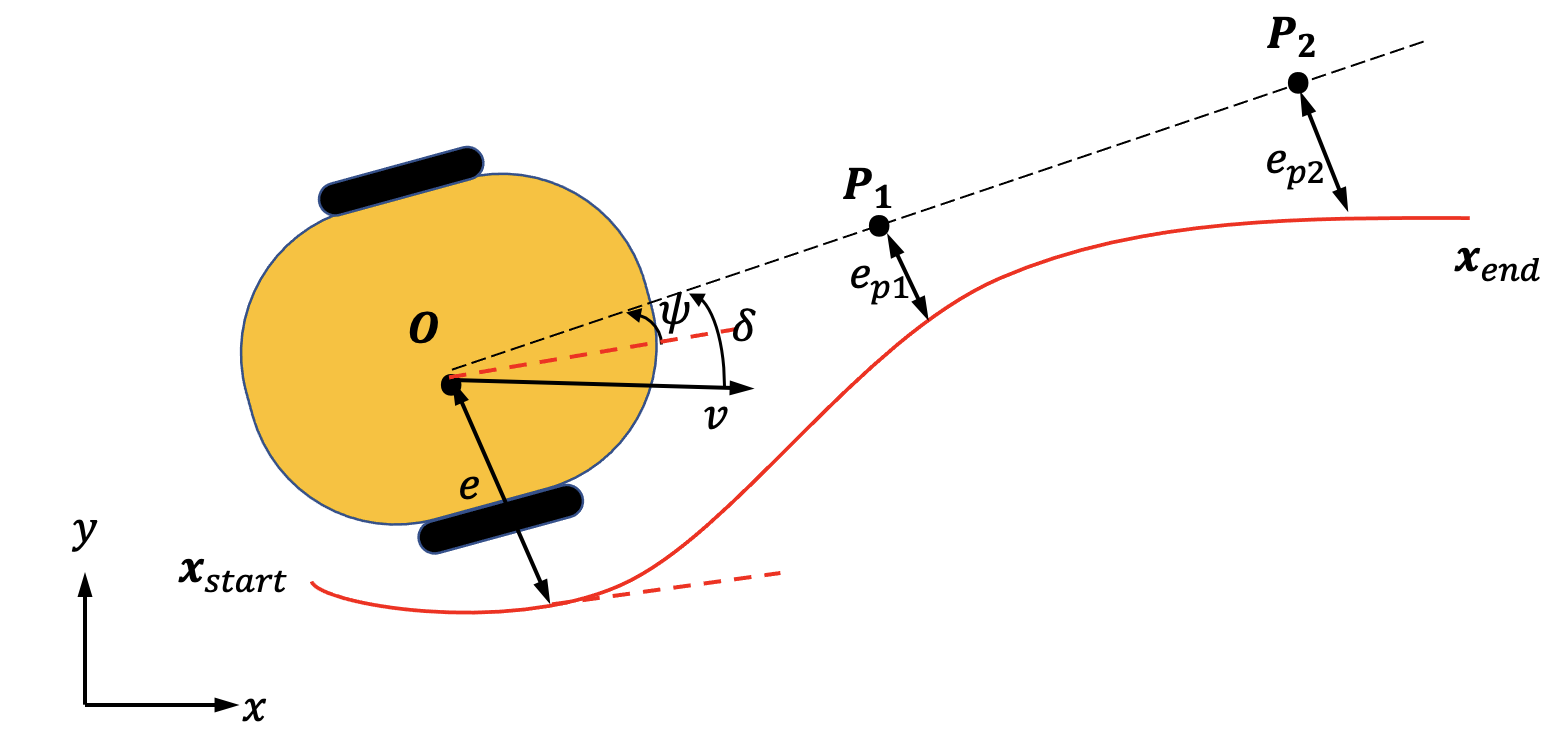}%
\caption{\textbf{Predictive Stanley Representation.}}
\label{fig:stanley_car_ref}
\end{figure}

 The proposed lateral control approach utilizes a version of Predictive Stanley controller, which is built on the basic Stanley controller. The basic Stanley controller is divided into three regions: saturated low region, saturated high region, and nominal region. $\psi$ is the heading of the vehicle with respect to the heading of the trajectory at the point of the projected shortest distance to the vehicle position, $e(t)$ is the lateral error, $v$ is current speed, and $K = K_{stanley}$ is the controller gain. See Fig. \ref{fig:stanley_car_ref} for reference. The steering angle command is given by:

\small
\begin{equation}
  \delta(t)=\begin{cases}
    \psi(t) + arctan(\frac{Ke(t)}{v(t)}), \hspace{2mm} |\psi(t) + arctan(\frac{Ke(t)}{v(t)})| < \delta(max)\\
    \delta(max), \hspace{9mm} \psi(t) + arctan(\frac{Ke(t)}{v(t)}) >= \delta(max) \\
    -\delta(max), \hspace{5mm} \psi(t) + arctan(\frac{Ke(t)}{v(t)}) <= -\delta(max)
  \end{cases}
\end{equation}
\normalsize

As discussed in \cite{abdelmoniem2020path}, the proposed predictive Stanley control approach introduces a third input, which is a developed array of future vehicle states, propagated along the vehicle track, denoted as $P_1$, $P_2$... $P_N$ as shown in Fig. \ref{fig:stanley_car_ref}. At each future state, the corresponding $\delta$ is calculated based on current $e_{pi}$. The final steering angle command is calculated by augmenting the output of each basic Stanley controller at each state to eliminate the error along the path not only at the reference point, as shown in Eqn. \ref{eqn:predictive_stanley_main} and \ref{eqn:predictive_stanley_p}. Consequently, the predictive Stanley controller is able to deal with the sudden changes in the heading angle of the trajectory by having this preview capability.

\begin{equation}
\delta(t)=\sum_{i=0}^{N} p_i [\psi_i(t) + arctan(\frac{Ke_{pi}(t)}{v(t)})]
\label{eqn:predictive_stanley_main}
\end{equation}

\begin{equation}
p_i = p_{i-1}^2 \hspace{2mm} \text{for} \hspace{2mm} i = 2...N
\label{eqn:predictive_stanley_p}
\end{equation}

\hspace{1cm} Eqn. 5-6. The final steering command. $p_i$ is the weight, which represents how each controller contributes in determining the final value of the steering angle. Here we set $N$ = 2 and $p_1$ = 0.2.

\begin{figure}[ht]
\centering
\includegraphics[width=0.8\linewidth]{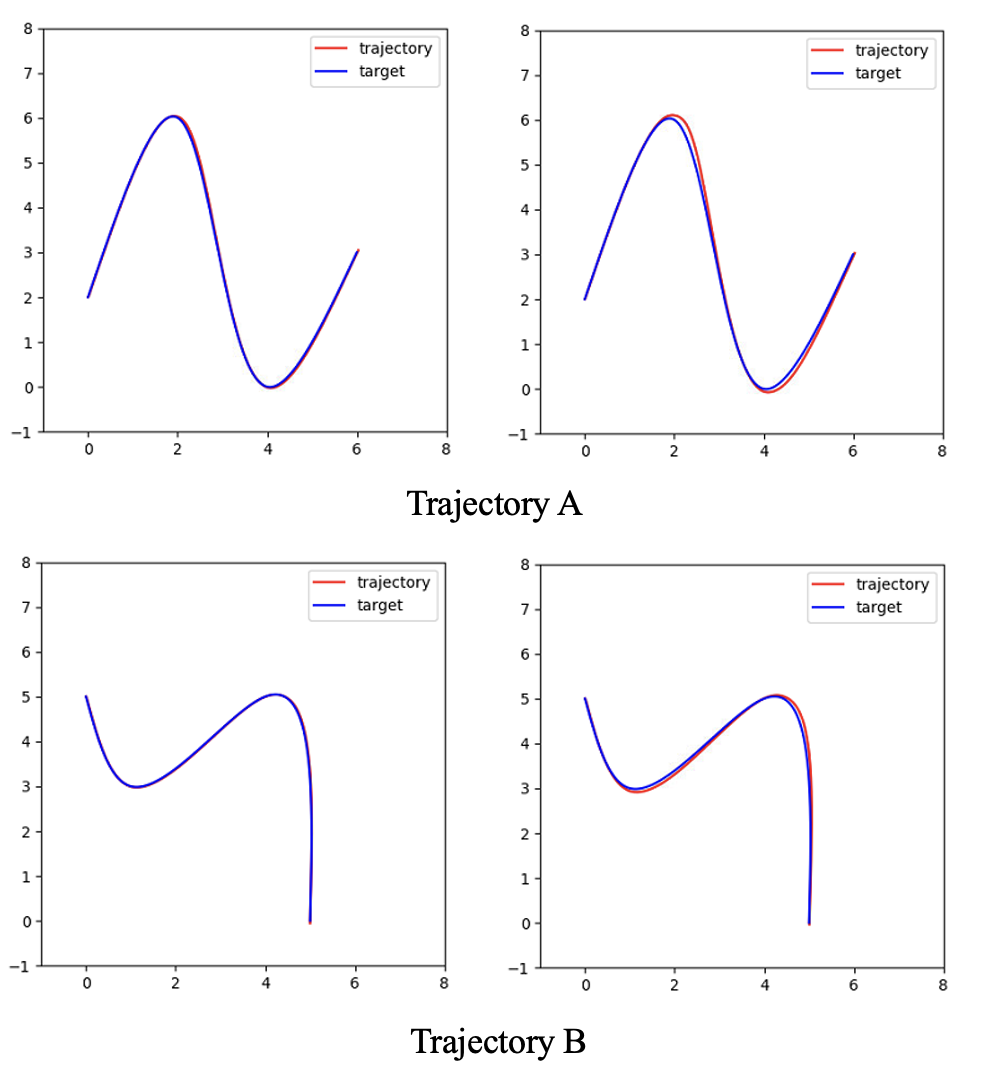}%
\caption{\textbf{Predictive Stanley vs. Basic Stanley Controller.} The average lateral error for trajectory A for predictive and basic Stanley controllers are $0.0147 m$ and $0.0374 m$, respectively; for trajectory B are $0.0077 m$ and $0.0347 m$, respectively.}
\label{fig:predictive_stanley_advantage}
\end{figure}

We show the advantage of our predictive Stanley controller over the basic Stanley controller by running them on a TurtleBot with the same trajectory. The visualization in Fig. \ref{fig:predictive_stanley_advantage} clearly shows that the predictive Stanley controller is able to adjust for abrupt turns. See detailed comparison in \cite{abdelmoniem2020path}.

\subsection{Longitudinal Control: PID}

We use a simple proportional control for speed regulation. The acceleration command becomes:

\begin{equation}
\alpha(t) = K_{speed}(v_{ref} - v(t))
\label{eqn:longitudinal_PID}
\end{equation}

With the steering and acceleration commands, we can deduce the robot's linear and angular velocity commands $[v, \omega]$ using Eqn. \ref{eqn:v_command} and \ref{eqn:w_command}, which are then executed by the low level controller.

\begin{equation}
v_{command} = v(t) + \alpha(t) \cdot \Delta T
\label{eqn:v_command}
\end{equation}

\begin{equation}
\omega_{command} = \frac{\delta(t)}{\Delta T}
\label{eqn:w_command}
\end{equation}

\subsection{Reinforcement Learning module}
The RL module in Fig. \ref{fig:framework} consists of actor and critic neural network layers. The entire framework in Fig. \ref{fig:framework} utilizes soft actor-critic (SAC) during training.

\section{Experiments}
Our experiments were designed and conducted in order to answer the following questions:
\begin{enumerate} 
\item Is simultaneous gain tuning necessary?

\item Is online gain tuning better than fixing the gains throughout the trajectory?

\item How to interpret our framework's output?

 \end{enumerate}

To answer these questions, we carry out simulated experiments using PyBullet by \cite{coumans2021}, with a TurtleBot waffle-pi model. To evaluate our framework, we propose to use a set of long-term and short-term metrics. Long-term metrics focus on measuring the performance throughout the entire trajectory, while short-term metrics focus on the short-time performance while the robot is slipping. Here we define slipping condition as those robot body states satisfying $\Delta v_{c\_vs\_actual} > |0.7| m/s$ or $\Delta \omega_{c\_vs\_actual} > |3| rad/s$.

For long-term criteria, we define average episodic reward $\overline r$, average lateral error $\overline {e}$, average speed error $\overline{\Delta v}$, and average RMS of change in low level control command $\overline{\Delta \textbf u}$(which measures command stability). $\mathbf{u} = [\omega_L, \omega_R]$, which denotes left and right wheel velocity control action commands, and is calculated based on $[v, \omega]$, using Eqn. \ref{eqn:v_transformation} and \ref{eqn:w_transformation}. The long-term metrics are calculated with respect to the entire trajectory.

\begin{equation}
\overline r = \frac{1}{T_{traj}} \sum_{i=0}^{T_{traj}}r_i
\label{eq_avg_errors}
\end{equation}

\begin{equation}
\overline {e} = \frac{1}{T_{traj}} \sum_{i=0}^{T_{traj}}e_i
\label{eq_avg_errors}
\end{equation}

\begin{equation}
\overline{\Delta v}= \frac{1}{T_{traj}} \sum_{i=0}^{T_{traj}}|v_i - v_{ref}|,
\label{eq_avg_errors}
\end{equation} 

\begin{equation}
\overline{\Delta \textbf u}= \frac{1}{T_{traj}} \sum_{i=1}^{T_{traj}} \lVert \textbf u_{i+1} - \textbf u_{i} \rVert ,
\label{eq_avg_errors}
\end{equation}

For short-term criteria, we define max lateral error throughout the trajectory $e_{max}$, average lateral error during slipping($T_{slip}$) $\overline{e}_{slip}$, average speed error during slipping $\overline{\Delta v}_{slip}$, and average RMS of change in low level  control action during slipping $\overline{\Delta \textbf u}_{slip}$.

\begin{equation}
e_{max} = max\{e_i\}_{0}^{T_{traj}}
\label{eq_avg_errors}
\end{equation}

\begin{equation}
\overline{e}_{slip} = \frac{1}{T_{slip}} \sum_{i=0}^{T_{slip}}e_i
\label{eq_avg_errors}
\end{equation}

\begin{equation}
\overline{\Delta v}_{slip}= \frac{1}{T_{slip}} \sum_{i=0}^{T_{slip}}|v_i - v_{ref}|,
\label{eq_avg_errors}
\end{equation} 

\begin{equation}
\overline{\Delta \textbf u}_{slip} = \frac{1}{T_{slip}} \sum_{i=1}^{T_{slip}} \lVert \textbf u_{i+1} - \textbf u_{i} \rVert ,
\label{eq_avg_errors}
\end{equation} 

\subsection{Simulation environment setup and training}
\begin{figure}[ht]
\centering
\includegraphics[width=1.0\linewidth]{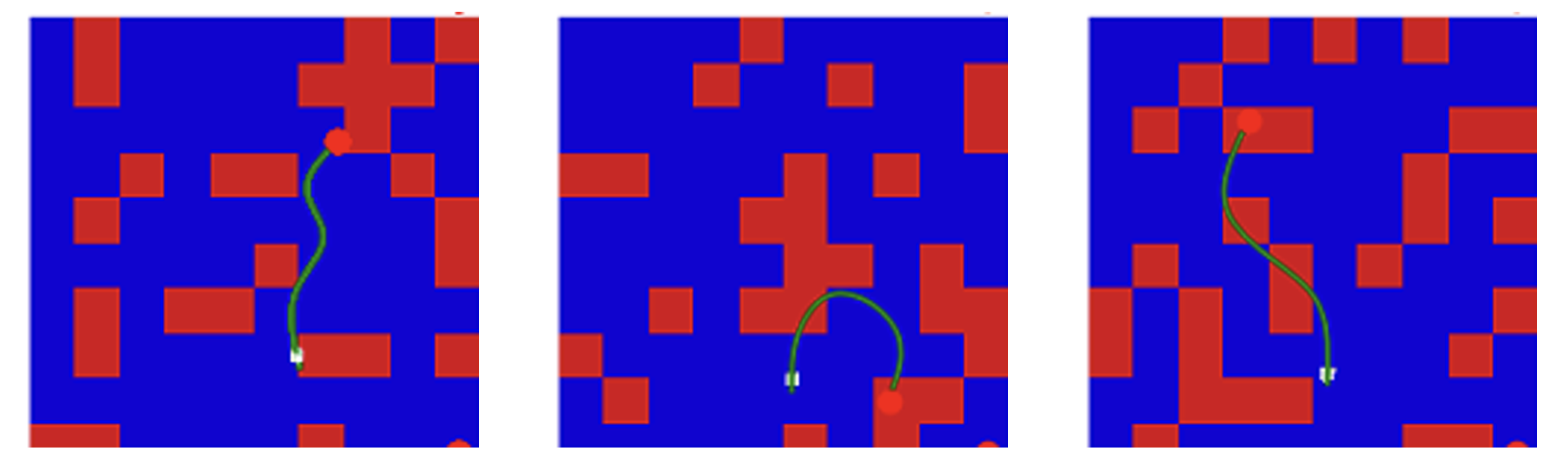}%
\caption{\textbf{Simulation Rendering}. Bird-eye view image with annotations. White dot is the robot, green line is the reference trajectory, and red dot is the goal position.}
\label{fig:simulation_rendering}
\end{figure}

Fig. \ref{fig:simulation_rendering} shows bird-eye view simulation renderings of three example setups. Blue color represents high frictional areas with frictional coefficient $\mu$=0.9, and red represents low frictional areas with $\mu$=0.01. The red patches are of size 1 $m$ by 1 $m$. The green trajectory is generated using a spline generator by \cite{sakai2018pythonrobotics}. The planner takes $n$ number of 2D points and generates a smooth trajectory connecting all the given points. 

To randomize the trajectory, we choose to use 5 random points for curve generation. Each point is $Uniform[1,2]$ $m$ away from the previous point, with $Uniform[-0.5\pi,0.5\pi]$ $rad$ angle from the previous point. The initial point $[x,y]$ position follows distribution: $x = Uniform[1,2]$ $m$, $y = Uniform[3.5,4.5]$ $m$.

To randomize the ground configuration, the red patches are randomly generated for each new trajectory, and we set 30\% of the total area to be red.

We use SAC algorithm to train the entire framework. The policy network and the value network in the SAC are fully-connected two-layer neural networks of size 64. Learning rate is 0.0006, $\gamma$ is set to be 0.99. The coefficients $R_{dist}$, $R_{ang}$ and $R_{speed}$ are set to be -20, -1, -1, respectively. The entire framework is trained until convergence.

\subsection{Is simultaneous gain tuning necessary?}

\begin{figure}[ht]
\centering
\includegraphics[width=0.9\linewidth]{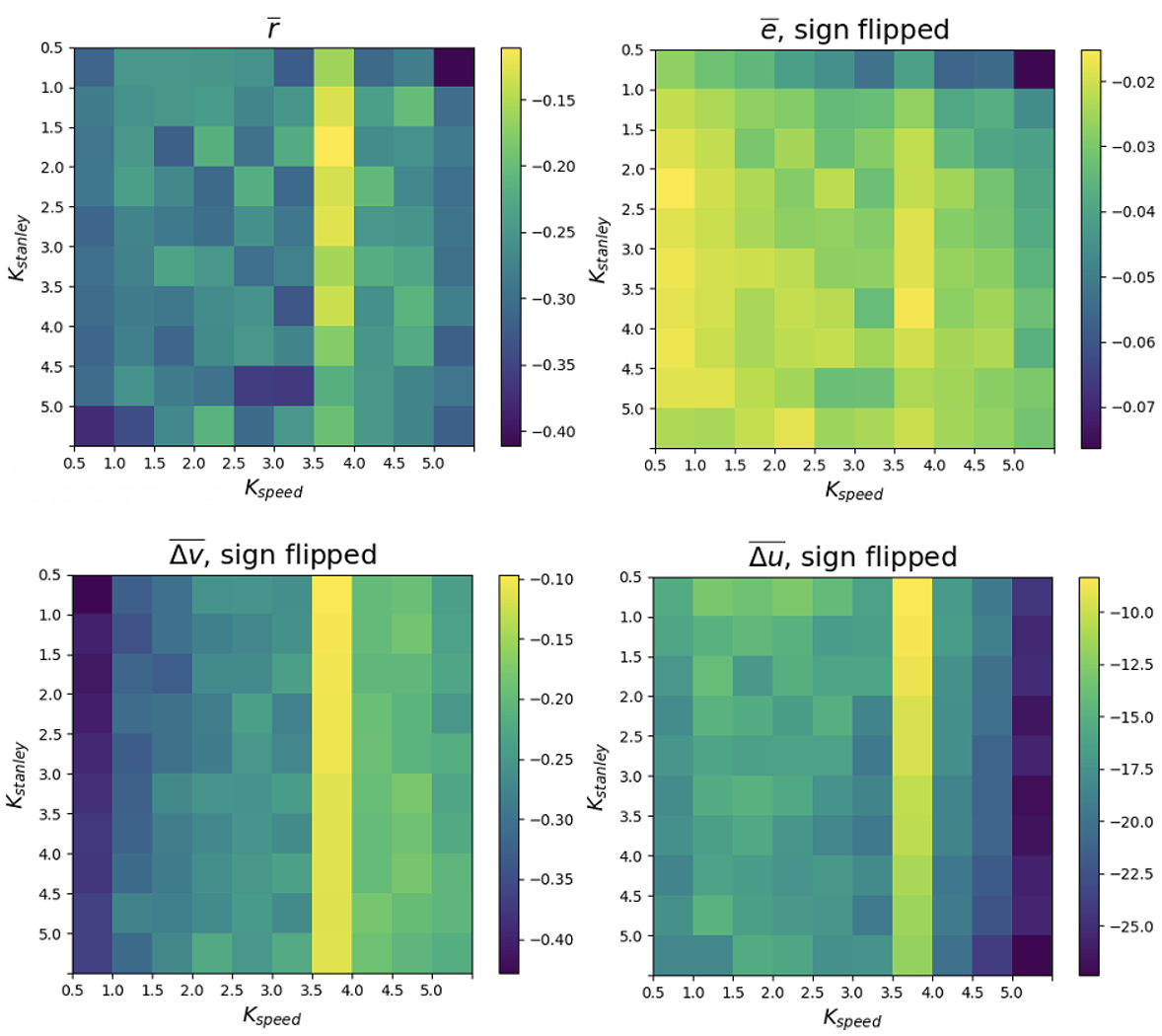}%
\caption{\textbf{Parameter sweeping results for long-term metrics} }
\label{fig:heatmap_long}
\end{figure}

\begin{figure}[ht]
\centering
\includegraphics[width=0.9\linewidth]{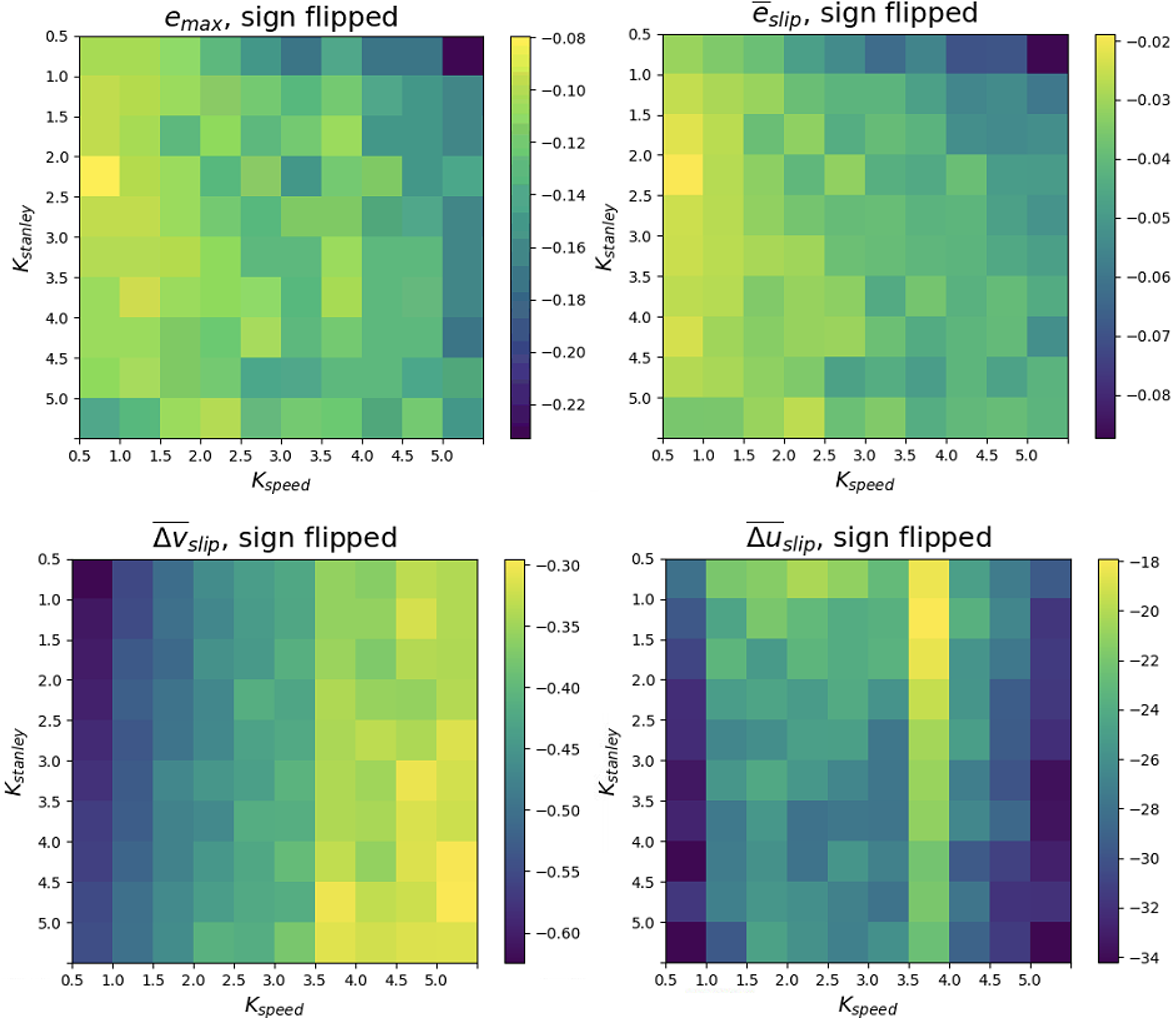}%
\caption{\textbf{Parameter sweeping results for short-term metrics} }
\label{fig:heatmap_short}
\end{figure}

We propose to utilize RL to tune the two gains simultaneously, rather than having two separate frameworks to determine each. To verify the necessity of simultaneous gain tuning, we vary $K_{stanley}$ and $K_{speed}$ from 0.5 to 5.0 with 0.5 increments, and plot heatmaps for each of the criteria discussed previously (Fig. \ref{fig:heatmap_long} and \ref{fig:heatmap_short}). Each point on the heatmap represents the result of using a specific combination of $K_{stanley}$ and $K_{speed}$. Each point result is calculated by averaging the results of running 100 pre-generated random trajectories with random ground setups. 

It can be shown that for each criteria, the best result happens when considering $K_{stanley}$ and $K_{speed}$ together. For example, for $\overline{e}$, fixing $K_{stanley}$ to be $2.5$ will result in a best $K_{speed}$ of $3.5$, but fixing $K_{stanley}$ to be $5.0$ will result in a best $K_{speed}$ of $2.0$. Therefore the gains have to be tuned simultaneously in order to achieve the best results.

For different criteria, the optimum happens at different combinations of $K_{stanley}$ and $K_{speed}$, because the criteria are focused on different aspects. For instance, to stabilize command and reduce $\overline{\Delta \mathbf{u}}$, $\overline{e}$ may be compromised because commands need to be tuned less abruptly.

\subsection{Is online gain tuning better than fixed parameters?}

We propose to tune the gains online throughout the entire trajectory rather than using fixed gains. To verify this, we use a baseline model. The baseline model uses the same framework as in Fig. \ref{fig:framework}, but without the RL module. Instead of varying the two gains online, the baseline model utilizes fixed best gains found by offline parameter sweeping in $K_{stanley}$ and $K_{speed}$. The best gain combinations of baseline model for each metric is shown in the second column in $Parameter\ Sweeping$ in Tables \ref{table:longterm_metric_comparison} and \ref{table:shortterm_metric_comparison}. We run the same 100 pre-generated random trajectories for the baseline model and our trained model, and log the results in the two tables. It can be shown that our framework is able to improve $\overline{e}$, $e_{max}$, $\overline{e}_{slip}$ by 6.6\%, 12.7\%, and 4.7\%, respectively.

One thing to notice is that the best results for each metric happens at different gain combinations with the baseline parameter-sweeping model. For example $\overline{e}$ has best results when setting $K_{stanley}=2.0$ and $K_{speed}=0.5$, but for $\overline{\Delta \mathbf{u}}$ it's $K_{stanley}=0.5$ and $K_{speed}=3.5$, which means the baseline model will perform worse if using the same combination of gains for all metrics. However our model is still able to beat the best of each baseline model metric with a universal trained policy, in lateral error metrics and $\overline{\Delta \mathbf{u}}$ metric.

\begin{table}[!ht]
    \centering
        \scriptsize
        \setlength{\tabcolsep}{0.2em} 
        \renewcommand{\arraystretch}{1.5}
        \caption{ \textbf{Long-term metrics comparison.} The second column in Parameter Sweeping indicates at what value of gains the best metric result was obtained.}
        \label{table:longterm_metric_comparison}
        \begin{tabular}{|c|c|c|c|c|}
        \hline
        Metrics & \multicolumn{2}{c|}{Parameter Sweeping} & Proposed Framework & Improvement\\
         \hline
         $(-)\overline{r}$ & $0.110\pm0.118$ & \vtop{\hbox{\strut $K_{stanley} = 1.5$}\hbox{\strut $K_{speed} = 3.5$}} & $0.084\pm0.125$ & 23.6\% \\
         \hline
         $\overline{e}$ & $0.015\pm0.010$ & \vtop{\hbox{\strut $K_{stanley} = 2.0$}\hbox{\strut $K_{speed} = 0.5$}} & $0.014\pm0.017$ & 6.6\% \\
        \hline
         $\overline{\Delta v}$ & $0.096\pm0.042$ & \vtop{\hbox{\strut $K_{stanley} = 0.5$}\hbox{\strut $K_{speed} = 3.5$}} & $0.097\pm0.058$ & -1.0\%\\
         \hline
        $\overline{\Delta \mathbf{u}}$ & $8.320\pm2.040$ & \vtop{\hbox{\strut $K_{stanley} = 0.5$}\hbox{\strut $K_{speed} = 3.5$}} & $5.917\pm2.165$ & 28.9\%\\
         \hline
    
        \end{tabular}
\end{table}

\begin{table}[!ht]
    \centering
        \scriptsize
        \setlength{\tabcolsep}{0.2em} 
        \renewcommand{\arraystretch}{1.5}
        \caption{ \textbf{Short-term metrics comparison.}}
        \begin{tabular}{|c|c|c|c|c|}
        \hline
        Metrics & \multicolumn{2}{c|}{Parameter Sweeping} & Proposed Framework & Improvement\\
         \hline
         $e_{max}$ & $0.079\pm0.055$ & \vtop{\hbox{\strut $K_{stanley} = 2.0$}\hbox{\strut $K_{speed} = 0.5$}} & $0.069\pm0.066$ & 12.7\% \\
         \hline
         $\overline{e}_{slip}$ & $0.021\pm0.013$ & \vtop{\hbox{\strut $K_{stanley} = 2.0$}\hbox{\strut $K_{speed} = 0.5$}} & $0.020\pm0.032$ & 4.7\% \\
        \hline
         $\overline{\Delta v}_{slip}$ & $0.295\pm0.082$ & \vtop{\hbox{\strut $K_{stanley} = 4.5$}\hbox{\strut $K_{speed} = 5.0$}} & $0.42\pm0.087$ & -42.3\%\\
         \hline
        $\overline{\Delta \textbf u}_{slip}$ & $18.9\pm5.93$ & \vtop{\hbox{\strut $K_{stanley} = 1.0$}\hbox{\strut $K_{speed} = 3.5$}} & $21.27\pm4.76$ & -12.5\%\\
         \hline
    
        \end{tabular}
        \label{table:shortterm_metric_comparison}
\end{table}

\subsection{Explainability of our framework output}

\begin{figure}[ht]
\centering
\includegraphics[width=1.0\linewidth]{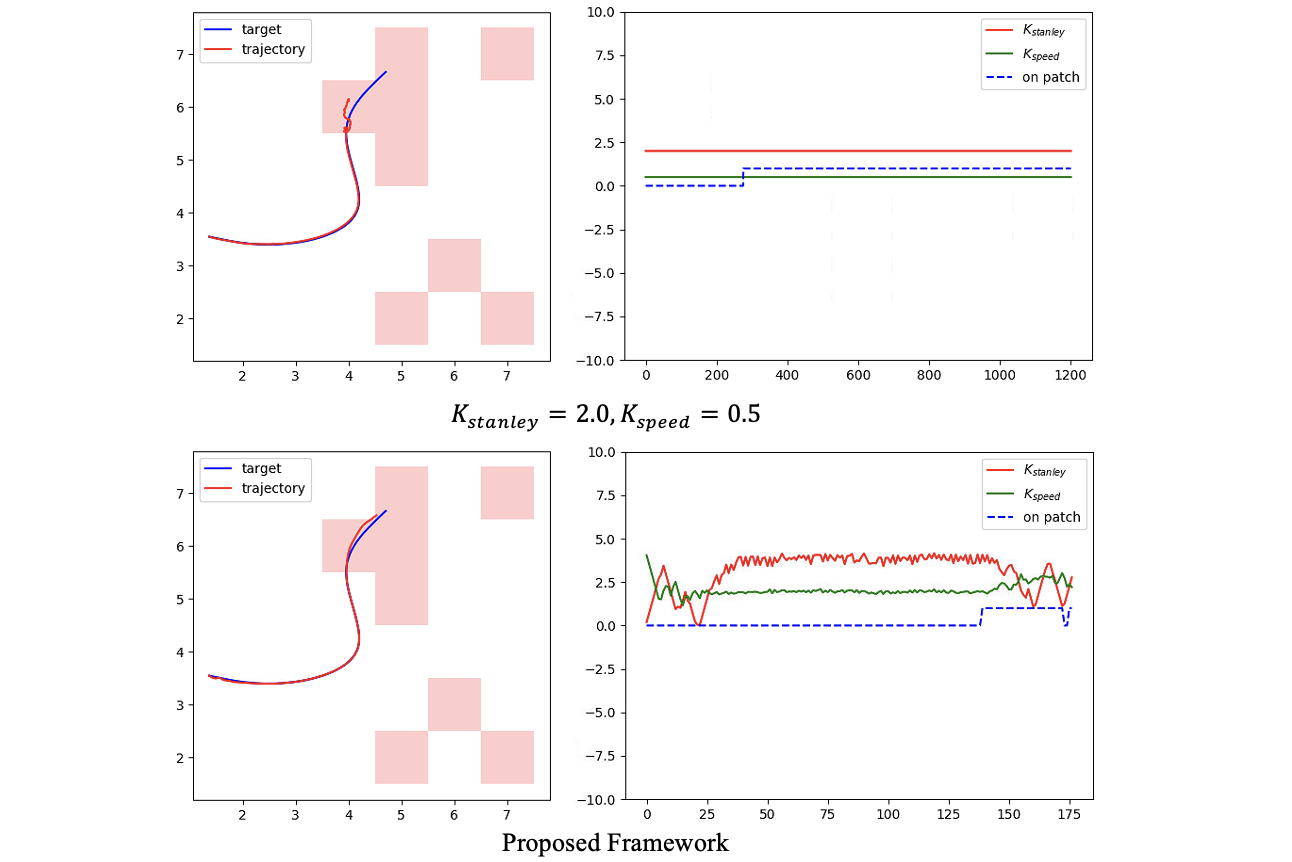}%
\caption{\textbf{Trajectory Comparison 1.} The left figures show the bird-view map. The red patches correspond to the low frictional area. The right figures show how the gains evolve versus time. The blue dotted line indicates when the robot is on the low frictional patch. A value of 1 indicates the robot is on patch.}
\label{fig:visualization_01}
\end{figure}

\begin{figure}[ht]
\centering
\includegraphics[width=1.0\linewidth]{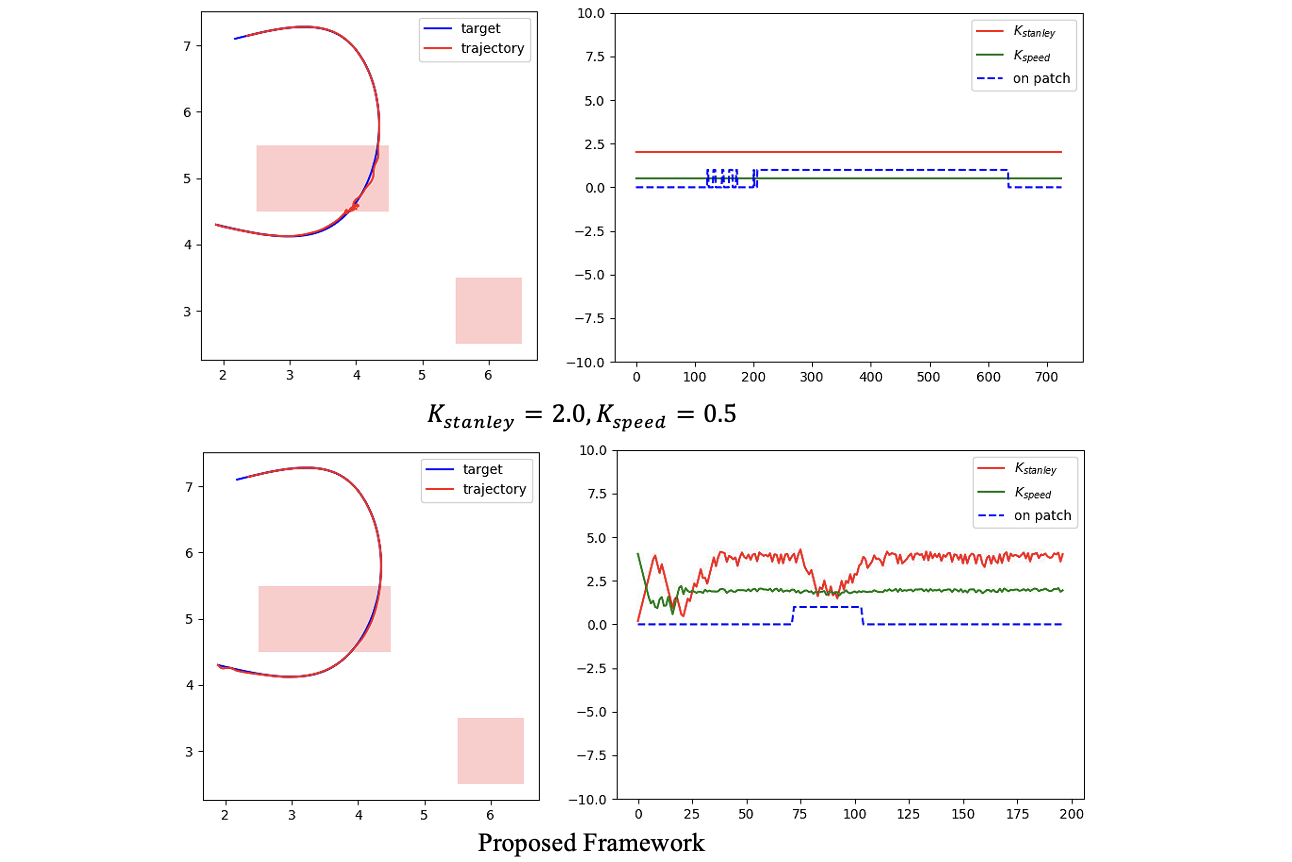}%
\caption{\textbf{Trajectory Comparison 2.} Using online tuning instead of fixing gains alleviates deviation when the robot enters the slippery area. $\overline{e}$ for baseline and proposed framework are $0.0247$ and $0.0112$, respectively. $e_{max}$ are $0.0624$ and $0.0336$, respectively.} 
\label{fig:visualization_02}
\end{figure}

We visualize two setup results in Fig. \ref{fig:visualization_01} and \ref{fig:visualization_02}. The upper figures show the baseline model with best gain combinations, and the lower figures show our model. 

In the first setup, the robot using the baseline model failed to reach the end and got stuck when first enters the patch area, while our model is able to succeed. A closer look in the right figure reveals that our model reduces $K_{stanley}$ when the robot enters the low frictional area and detects slip. It makes sense as when slip happens, heavy steering will not help much and could worsen the slip. A good tuning on $K_{stanley}$ and $K_{speed}$ helps the robot to control the slip and succeed in the tracking task in this case. And it is up to the RL module to decide when and how much to tune the two gains. Also, the RL module does not have prior knowledge of the location of low-frictional area, and it is able to tune the gains based on current tracking status and reduce lateral error successfully.

Similar observation can be made in the second setup. Both models succeeded in the task, but the baseline model induced more lateral error when the robot entered low-frictional area, because it did not lower $K_{stanley}$ accordingly.

\section{Discussion}

The experiments conducted show that simultaneous and online tuning of gains are necessary for mobile robot trajectory tracking under slippery conditions. Our model is able to tune the gains in lateral and longitudinal controls, and beat the baseline model in terms of lateral error metrics. 

To reason the need of simultaneous gain tuning, consider when the robot is slipping or trajectory contains a large curvature. Both acceleration and steering commands should be considered to achieve an optimal tracking performance. For instance, with a large curvature, speed regulation can be relaxed while the steering command needs to have a bigger gain. Or when robot is slipping, both gains might need to be adjusted, as seen in Fig. \ref{fig:visualization_01} and \ref{fig:visualization_02}. The magnitude of commands needs to be determined simultaneously, and the RL module in our framework decides the magnitudes of both gains at each timestep.

Also notice for some metrics such as $\overline{\Delta \textbf u}$ and $\overline{\Delta \textbf u}_{slip}$, $K_{speed}$ tuning dominates the performance. One reason is that speed regulation impacts the wheel velocity change more than angular regulation, because sometimes the curvature is not very abrupt for steering to change much, but speed has to be regulated all the time.

 We showed our framework is able to improve $\overline{e}$, $e_{max}$, $\overline{e}_{slip}$ by 6.6\%, 12.7\%, and 4.7\%, respectively. However the command and speed stability during slipping were compromised a bit. It makes sense as the robot tries to relax other constraints in order to reduce the lateral error. In many mobile robot slipping scenarios such as in a factory setting, the most important aspect is to reduce the lateral error because deviations could lead to robot hitting unwanted objects and causing harms. Therefore a lower stability in speed and command are acceptable.


\section{Conclusion and Future Work}

To reduce slip for mobile robots, we propose a hierarchical framework that utilizes an RL module to adapt gains for the tracking controllers simultaneously online. We demonstrated the necessity of simultaneous gain tuning, and showed that our online framework outperforms the best baseline model using fixed gains, especially in terms of long and short term lateral errors.




\bibliography{references}

\end{document}